\documentclass[conference]{IEEEtran}
\IEEEoverridecommandlockouts
\usepackage{multirow}

\usepackage{graphicx}
\usepackage[tight,footnotesize]{subfigure}
\usepackage{color}
\usepackage{array}
\usepackage{tabularx}
\usepackage[strict]{changepage}
\usepackage{amssymb}
\usepackage{pifont}
\usepackage{enumitem}
\usepackage{gensymb}
\usepackage{makecell}
\usepackage{textcomp}
\usepackage{float}
\usepackage{balance}
\usepackage{mathtools}
\usepackage{ctable} 
\usepackage{amsthm}

\usepackage{algorithm}
\usepackage{algorithmic}
\usepackage{arydshln}
\DeclarePairedDelimiter\ceil{\lceil}{\rceil}
\DeclarePairedDelimiter\floor{\lfloor}{\rfloor}
\bstctlcite{IEEEexample:BSTcontrol}

\newcolumntype{L}[1]{>{\raggedright\let\newline\\\arraybackslash\hspace{0pt}}m{#1}}
\newcolumntype{C}[1]{>{\centering\let\newline\\\arraybackslash\hspace{0pt}}m{#1}}
\newcolumntype{R}[1]{>{\raggedleft\let\newline\\\arraybackslash\hspace{0pt}}m{#1}}

\usepackage[numbers]{natbib}

\def\tsc#1{\csdef{#1}{\textsc{\lowercase{#1}}\xspace}}
\tsc{WGM}
\tsc{QE}

\begin{document}
\let\WriteBookmarks\relax
\def\floatpagepagefraction{1}
\def\textpagefraction{.001}




\title{AMLNet: Adversarial Mutual Learning Neural Network for Non-AutoRegressive Multi-Horizon Time Series Forecasting
}


\author{\IEEEauthorblockN{Yang Lin}
	\IEEEauthorblockA{\textit{School of Computer Science, The University of Sydney}\\
		Sydney, Australia \\
		ylin4015@uni.sydney.edu.au}
}

\maketitle

\begin{abstract}
Multi-horizon time series forecasting, crucial across diverse domains, demands high accuracy and speed. While AutoRegressive (AR) models excel in short-term predictions, they suffer speed and error issues as the horizon extends. Non-AutoRegressive (NAR) models suit long-term predictions but struggle with interdependence, yielding unrealistic results. We introduce AMLNet, an innovative NAR model that achieves realistic forecasts through an online Knowledge Distillation (KD) approach. AMLNet harnesses the strengths of both AR and NAR models by training a deep AR decoder and a deep NAR decoder in a collaborative manner, serving as ensemble teachers that impart knowledge to a shallower NAR decoder. This knowledge transfer is facilitated through two key mechanisms: 1) outcome-driven KD, which dynamically weights the contribution of KD losses from the teacher models, enabling the shallow NAR decoder to incorporate the ensemble's diversity; and 2) hint-driven KD, which employs adversarial training to extract valuable insights from the model's hidden states for distillation. Extensive experimentation showcases AMLNet's superiority over conventional AR and NAR models, thereby presenting a promising avenue for multi-horizon time series forecasting that enhances accuracy and expedites computation.
\end{abstract}




\maketitle


\section{Introduction}

Time-series forecasting is integral to various practical applications, from electricity grid control \cite{9653868} and economic trend prediction \cite{doi:10.1137/1.9781611976700.84} to traffic control \cite{pmlr-v162-zhou22g}. Such scenarios often require multi-step ahead forecasts for informed decision-making; for instance, accurately predicting hourly electricity consumption for upcoming days or weeks aids efficient resource allocation. Classical statistical methods like AutoRegressive Integrated Moving Average (ARIMA) and exponential smoothing \cite{Durbin01book} excel in single time series forecasting, yet fall short when handling related time series collectively, as they treat each series independently \cite{AST20NIPS,N-BEATS,Logsparse19NIPS}. Emerging as a promising alternative, deep learning techniques have gained traction for large-scale related time series forecasting \cite{DeepAR20,Logsparse19NIPS,9679135,pmlr-v162-zhou22g}.

These deep learning methods can be categorized into AutoRegressive (AR) and Non-AutoRegressive (NAR) models. AR models, including DeepAR \cite{DeepAR20}, TCNN \cite{TCNN_Yang,TCAN_Yang} and LogSparse Transformer \cite{Logsparse19NIPS}, predict one step ahead, using prior forecasts as input for subsequent predictions. While effective at capturing interdependencies in the output space \cite{Gu18ICLR,barezi2020study}, AR models face issues such as training-inference discrepancies \cite{bengio2015scheduled,MQRNN18}, error accumulation \cite{Taieb16}, and high inference latency \cite{Cho16,Lee18EMNLP}. Conversely, NAR models (e.g., MQ-RNN \cite{MQRNN18}, N-BEATS \cite{N-BEATS}, AST \cite{AST20NIPS}, and Informer \cite{Informer21}) overcome AR modeling problems by generating parallel predictions, proving superior in long horizon forecasting. However, NAR models may yield unrealistic, disjointed series due to a lack of interdependence consideration, leading to unrelated forecasts \cite{Taieb16,AST20NIPS}. Our work addresses this by employing Knowledge Distillation (KD) to incorporate both model outcomes and hidden states, yielding more coherent and accurate forecasts.

To tackle NAR model limitations, we introduce Adversarial Mutual Learning Neural Network (AMLNet), an NAR model utilizing online KD methods. AMLNet comprises an encoder, a deep AR decoder, a deep NAR decoder, and a shallow NAR decoder (Fig. \ref{AMLNet}). During training, encoder extracts patterns for all decoders, deep AR and NAR decoders train in mutual fashion, then serve as ensemble teachers transferring knowledge to the shallow NAR decoder, enhancing error handling and output interdependence. Testing employs the encoder and shallow NAR for forecast generation. AMLNet's knowledge transfer employs two techniques: outcome-driven KD, dynamically weighting distillation loss based on network performance to prevent error circulation; and hint-driven KD, distilling knowledge from hidden states via adversarial training, as these states contain valuable information for enhanced transfer.

Our contributions encompass:
1) Introduction of AMLNet, pioneering online KD for time series forecasting. It trains deep AR and NAR decoders mutually as ensemble teachers, transferring knowledge to a shallow NAR decoder, resulting in contiguous forecasts and superior performance with fast inference speed, as demonstrated across four time series datasets.
2) Proposal of outcome-driven and hint-driven online KD, simultaneously learning from teacher network predictions and inner features. Our experiments, compared to state-of-the-art KD methods, affirm the efficacy of both proposed techniques.

\section{Problem Formulation}

\subsection{Data Sets}
We conducted experiments on four publicly available real-world datasets: Sanyo \cite{San}, Hanergy \cite{Han}, Solar \cite{Sol}, and Electricity \cite{Ele}.
The \textbf{Sanyo} and \textbf{Hanergy} datasets consist of solar power generation data from two PV plants in Australia. The data for Hanergy spans from 01/01/2011 to 31/12/2016 (6 years), while the data for Sanyo covers the period from 01/01/2011 to 31/12/2017 (7 years).
The \textbf{Solar} dataset comprises solar power data from 137 PV plants in Alabama, USA, gathered between 01/01/2006 and 31/08/2006.
The \textbf{Electricity} dataset contains electricity consumption data from 370 households, recorded from 01/01/2011 to 07/09/2014.

A summary of the data statistics is provided in Table \ref{datasets}. For the Sanyo and Hanergy datasets, we considered data between 7 am and 5 pm and aggregated it at half-hourly intervals. Additionally, weather and weather forecast data were collected and used as covariates in the experiments (refer to \cite{Yang20ICONIP} for further details). 
For the Solar and Electricity datasets, the data was aggregated into 1-hour intervals. Following the approach in \cite{Logsparse19NIPS,Yang20ICONIP}, calendar features were incorporated based on the granularity of each dataset. Specifically, Sanyo and Hanergy datasets used features such as \textit{month, hour-of-the-day, and minute-of-the-hour}, Solar dataset used \textit{month, hour-of-the-day, and age}, and Electricity dataset used \textit{month, day-of-the-week, hour-of-the-day, and age}. 
For consistent preprocessing, all data was normalized to have zero mean and unit variance \cite{Yang20ICONIP}.

\begin{table*}[!h]
	\renewcommand{\arraystretch}{1}	
	\centering
	\begin{tabular}{C{1.5cm} C{1.8cm} C{1.8cm} C{2cm} C{.8cm} C{.8cm} C{.8cm} C{.8cm} C{.8cm} C{.8cm} } 
		\specialrule{.1em}{.05em}{.05em} 
		&Start date&End date &Granularity &$L_d$&$N$& $n_{T}$  & $n_{C}$& $T_{l}$  & $T_{h}$ \\	
		\hline
		Sanyo&01/01/2011& 31/12/2016 & 30 minutes&20 & 1 &4 &3&20 &20\\				
		Hanergy&01/01/2011& 31/12/2017 & 30 minutes&20 & 1 &4 &3&20 &20\\			
		Solar&01/01/2006& 31/08/2006 & 1 hour&24 & 137 &0 &3&24 &24\\			
		Electricity&01/01/2011& 07/09/2014 & 1 hour&24 & 370 &0 &4&168 &24\\		
		\specialrule{.1em}{.05em}{.05em} 
	\end{tabular}
	\caption{Dataset statistics. $L_d$ - number of steps per day, $N$ - number of series, $n_{T}$ - number of time-based features, $n_{C}$ - number of calendar features, $T_{l}$ - length of input series, $T_{h}$ - length of forecasting horizon. }
	\label{datasets} 
\end{table*}

\subsection{Problem Statement}

Given is: 1) a set of $N$ univariate time series (solar or electricity series) $\{\mathbf{Y}_{i,1:T_l}\}^N_{i=1}$, where $\mathbf{Y}_{i,1:T_l}\coloneqq[{y}_{i,1}, {y}_{i,2},...,{y}_{i,T_l}]$, $T_l$ is the input sequence length, and ${y}_{i,t}\in\Re$ is the value of the $i$th time series (generated PV solar power or consumed electricity) at time $t$; 2) a set of associated time-based multi-dimensional covariate vectors $\{\mathbf{X}_{i, 1: T_l+T_h}\}_{i=1}^{N}$, where $T_h$ denotes the length of the forecasting horizon. 
Our goal is to predict the future values of the time series $\{\mathbf{Y}_{i,T_l+1:T_l+T_h}\}^N_{i=1}$, i.e. the PV power or electricity usage for the next $T_h$ time steps after $T_l$.

The covariates for the Sanyo and Hanergy datasets include: weather data $\{\mathbf{W1}_{i, 1: T_l}\}_{i=1}^{N}$, weather forecasts $\{\mathbf{WF}_{i, T_l+1: T_l+T_h}\}_{i=1}^{N}$ and calendar features $\{\mathbf{Z}_{i, 1: T_l+T_h}\}_{i=1}^{N}$, while the covariates for Solar and Electricity datasets include only calendar features.

Specifically, AMLNet produces the probability distribution of the future values, given the past history:
\begin{equation}
	\begin{split}
		p\left(\mathbf{Y}_{i,T_l+1:T_l+T_h} \mid \mathbf{Y}_{i, 1: T_l}, \mathbf{X}_{i, 1: T_l+T_h} ; \theta\right)\\ = \prod_{t=T_{l}+1}^{T_{l}+T_{h}} p\left({y}_{i, t} \mid \mathbf{Y}_{i, 1: T_l},  \mathbf{X}_{i, 1: T_l+T_h} ; \theta\right),
	\end{split}
\end{equation}
where the input of model at step $t$ is the concatenation of ${y}_{i, t-1}$ and ${x}_{i, t}$.

\section{Related Work}

\subsection{Non-AutoRegressive Sequence Modelling}	

NAR  forecasting models \cite{MQRNN18,N-BEATS,AST20NIPS,Informer21} directly eliminate AR connection from the decoder side, instead modeling separate conditional distributions for each prediction independently. Unlike AR models, NAR models enable parallelized training and inference processes. However, NAR models can produce disjointed forecasts and introduce discontinuities \cite{Taieb16} due to the erroneous assumption of independence, limiting their ability to capture interdependencies among predictions. AST \cite{AST20NIPS} stands as the sole approach addressing this within the NAR framework, utilizing adversarial training to enhance global perspective. 
Recent NAR models focused on reducing output space interdependence have primarily emerged in the realm of Natural Language Processing (NLP) tasks \cite{Gu18ICLR,Ren19NIPS}. Various strategies have emerged, with KD \cite{KD15,Kim16EMNLP} garnering substantial attention. KD effectively transfers knowledge from a larger teacher to a smaller student network by offering softer, more informative target distributions. KD methods for NAR either distill the prediction distribution of a pre-trained AR teacher model \cite{Gu18ICLR} or incorporate hidden state patterns of the AR model \cite{Hint19EMNLP}.

\subsection{Online Knowledge Distillation}	
Classic KD methods are offline and can incur computational and memory overhead due to the reliance on powerful pre-trained teachers. In response, online KD techniques \cite{DML18CVPR,Chung20ICML,AMLN20ECCV,Wu21AAAI} have emerged, showing superior results. These methods treat all networks as peers, enabling mutual exchange of output information and requiring less training time and memory. DML \cite{DML18CVPR} introduced collaborative training, where each model can be both a student and a teacher. Further advancements, such as Wu and Gong's work \cite{Wu21AAAI}, assemble teachers into online KD, enhancing generalization. Notably, online KD techniques can capture intermediate feature distributions through adversarial training, as seen in AFD \cite{Chung20ICML} and AMLN \cite{AMLN20ECCV}.

\subsection{Generative Adversarial Networks}	
Generative Adversarial Networks (GANs) \cite{GAN14NIPS}, comprising a generator $G$ and a discriminator $D$ engaged in adversarial training, were initially proposed for sample generation. However, the adversarial training paradigm has found applications in diverse domains, including computer vision, NLP \cite{Chung20ICML,AMLN20ECCV}, and time series forecasting \cite{GAN20ICML,AST20NIPS}.

\subsection{Summary}	
In contrast to prior work, our AMLNet introduces several advancements:
1) We pioneer the application of online KD for forecasting, introducing AMLNet as the first model to employ online KD methods for training a NAR forecasting model to capture target sequence interdependencies. Specifically, AMLNet trains a deep AR and a deep NAR model mutually as ensemble teachers before transferring their knowledge to a shallow NAR student.
2) While Wu and Gong \cite{Wu21AAAI} construct ensemble teachers and adjust KD loss weights based on training epoch number, they overlook teacher model instability during training. AMLNet utilizes outcome-driven KD, assigning dynamic weights to KD losses based on teacher model performance, specifically tailored for probabilistic forecasting.
3) We address the issue of discontinuous predictions stemming from NAR model hidden states, proposing hint-driven KD to capture hidden state distribution information. Unlike previous approaches \cite{Chung20ICML,AMLN20ECCV}, designed for networks with differing layer counts, our method is tailored to AMLNet's architecture.

\section{Adversarial Mutual Learning Neural Network}
The proposed architecture, Adversarial Mutual Learning Neural Network (AMLNet), addresses the challenges in NAR forecasting by modeling output space interdependence. 
This section presents the architecture of AMLNet, the proposed outcome-driven and hint-driven online KD methods and the optimisation and inference process of AMLNet. 

\begin{figure*}[!ht]
	\centering	  
	\includegraphics[width=1.7\columnwidth]{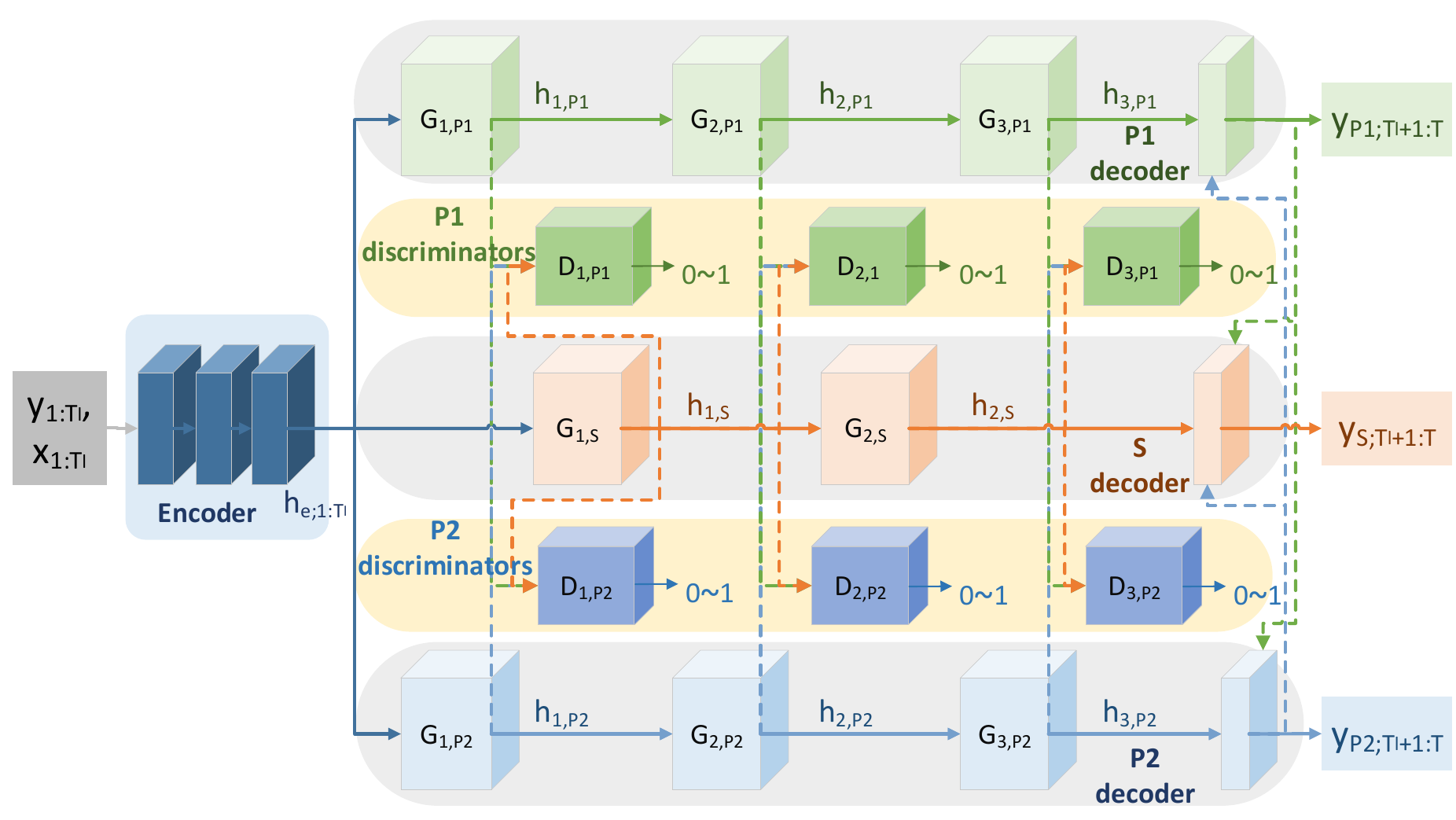}
	\caption{AMLNet comprises an encoder, P1, P2, and S decoders, with each P1 and P2 layer accompanied by a dedicated discriminator. P1 operates as an AR component, while P2 and S function as NAR components. The solid lines depict the feedforward process, while dashed lines represent the data flow for knowledge distillation.
	}
	\label{AMLNet}
\end{figure*}

\subsection{Network Architecture}
The central components of AMLNet, as depicted in Figure 1, encompass a shared encoder $f_{\theta_e}$ consisting of $n_e$ layers, a deep AR Peer1 (P1) decoder $f_{\theta_{P1}}$ with $n_P$ hidden layers, a deep NAR Peer2 (P2) decoder $f_{\theta_{P2}}$ also possessing $n_P$ hidden layers, and a shallow NAR Student (S) decoder $f_{\theta_{S}}$ equipped with $n_{S}$ hidden layers. It's noteworthy that Informer \cite{Informer21} serves as the foundational framework for AMLNet, although other deep learning forecasting models can replace Informer.

To harness temporal patterns from historical data, the encoder extracts insights from past time steps by processing the concatenation of covariate $x_t$ and ground truth $y_{t}$ as input at time step $t$:
\begin{equation}
	h_{e;1:T_l}=f_{\theta_e}(y_{1:T_l},x_{1:T_l})
	\label{theta_e}
\end{equation}
The shared encoder temporal patterns $h_{e;1:T_l}$ are uniformly leveraged across all decoders, exploiting the consistent input pasting sequence. This shared approach significantly reduces network parameters and computational overhead.

The P1, P2 and S decoders are formulated:
\begin{equation}
	y_{P1;T_l+1:T}=f_{\theta_{P1}}(y_{T_l:T-1},x_{T_l+1:T},h_{e;1:T_l})
	\label{theta_P1}
\end{equation}
\begin{equation}
	y_{P2;T_l+1:T}=f_{\theta_{P2}}(x_{T_l-T_de:T_l},h_{e;1:T_l})
	\label{theta_P2}
\end{equation}
\begin{equation}
	y_{S;T_l+1:T}=f_{\theta_{S}}(x_{T_l-T_de:T_l},h_{e;1:T_l})
	\label{theta_S}
\end{equation}
where $T_de$ is the length of start token used by Informer \cite{Informer21}, and the prediction $y$ consists of a mean and variance. NAR models without input sequences for the decoder have poor performance and copying input pasting series to the decoder as its input could enhance model performance \cite{Gu18ICLR}.

Notably, the AR model excels at capturing output space interdependence, whereas the NAR model is adept at mitigating error propagation. Through a mutually beneficial relationship, the deep AR P1 and NAR P2 decoders coalesce as peers, leveraging each other's strengths to augment their individual abilities. This collective knowledge is then harnessed to train the shallow NAR S decoder, effectively establishing a dynamic ensemble, as illustrated in Figure \ref{AMLNet}.

\subsection{Outcome-driven Online Knowledge Distillation}

Conventional optimization objective of probabilistic time series forecasting is the Negative Log Likelihood (NLL) loss, denoted as $\mathcal{L_{NLL}}$. This loss function, prevalent in training probabilistic forecasting models \cite{DeepAR20,Logsparse19NIPS}, is formally defined in Eq. (\ref{NLL}):

\begin{equation}
	\begin{split}
		&\mathcal{L_{NLL}}(\hat{y}_{T_l+1:T},y_{T_l+1:T})\\
		=&-\frac{1}{2T_h}\times\Big(T_h \log (2 \pi)+ \sum_{t=T_l+1}^{T} \log \left|\sigma_t^2\right| \\
		&+\sum_{t=T_l+1}^{T} (y_t-{\mu}_t)^{2} \sigma_{t}^{-2} \Big)
	\end{split}
	\label{NLL}
\end{equation}
where $y_{T_l+1:T}$ represents the ground truth, while $\hat{y}_{T_l+1:T}$ pertains to the predicted distribution, encompassing the mean ${\mu}t$ and standard deviation $\sigma{t}$ across the forecasting horizon.

Addressing the inherent limitations of both autoregressive (AR) and non-autoregressive (NAR) models when subjected to the NLL loss, we propose a novel approach by enlisting the counterpart model as a peer network. This peer relationship serves as a guiding mechanism to overcome the respective limitations, approximating both the ground truth and the predicted distribution of their teacher network.

In the realm of online Knowledge Distillation (KD), traditional methodologies involve aggregating prediction and KD losses with fixed predefined proportions \cite{DML18CVPR} or gradually increasing the proportion of KD loss over training epochs \cite{Wu21AAAI}. However, these approaches overlook the diverse abilities of the peer networks and the varying quality of their predictions during training. Consequently, the contribution of each peer network to the student network should be weighted according to its performance.

By allocating a constant weight to the KD loss irrespective of the peers' performance, inaccurately predicted distributions could propagate errors within the network, limiting overall performance. Conventional remedies, such as removing misclassified data for offline KD, are unsuitable for online distillation or forecasting tasks due to reduced training data.

To address these challenges, we introduce an attention-based KD loss that assigns higher weights to well-forecasted samples. The KD loss functions for the P1, P2, and S decoders are formulated in Eq. (\ref{P1}), (\ref{P2}), and (\ref{S}). P1 and P2 aim to mimic each other's predicted distributions, while S simultaneously distills knowledge from the outputs of both P1 and P2. The Kullback Leibler (KL) divergence serves as a measure of discrepancy between predicted distributions. Given distributions $P$ and $Q$, their KL divergence is defined in Eq. (\ref{KLdivergence}).
\begin{equation}
	\mathcal{L_{\mathrm{KL}}}(P \| Q)=\int_{-\infty}^{\infty} p(x) \log \left(\frac{p(x)}{q(x)}\right) d x
	\label{KLdivergence}
\end{equation}

\begin{equation}
	\begin{split}
		\mathcal{L}_{P1}=
		& \frac{\alpha_o}{T_h}\times \omega_e(y_{P2;T_l+1:T},y_{T_l+1:T})\cdot \\&\mathcal{L_{\mathrm{KL}}}(y_{P2;T_l+1:T}\|y_{P1;T_l+1:T})
	\end{split}
	\label{P1}
\end{equation}
\begin{equation}
	\begin{split}
		\mathcal{L}_{P2}=
		& \frac{\alpha_o}{T_h}\times \omega_e(y_{P1;T_l+1:T},y_{T_l+1:T})\cdot \\&\mathcal{L_{\mathrm{KL}}}(y_{P1;T_l+1:T}\|y_{P2;T_l+1:T})
	\end{split}
	\label{P2}
\end{equation}
\begin{equation}
	\begin{split}
		\mathcal{L}_{S}=
		& \frac{\alpha_o}{T_h}\times \omega_e(y_{P1;T_l+1:T},y_{T_l+1:T})\cdot \\&\mathcal{L_{\mathrm{KL}}}(y_{P1;T_l+1:T}\|y_{S;T_l+1:T})+
		\\& \frac{\alpha_o}{T_h}\times \omega_e(y_{P2;T_l+1:T},y_{T_l+1:T})\cdot  \\&\mathcal{L_{\mathrm{KL}}}(y_{P2;T_l+1:T}\|y_{S;T_l+1:T})
	\end{split}
	\label{S}
\end{equation}

In these equations, $\alpha_o$ modulates the weight of the outcome-driven KD loss, while the weight $\omega_e(\cdot)$ captures the importance of the teacher model's predictions. Considering a Gaussian distribution of data, we define $\omega_e(\cdot)$ as follows:
\begin{equation}
	\begin{split}
		&\omega_e(\hat{y}_{T_l+1:T},y_{T_l+1:T})
		=\frac{1}{\sigma_{T_l+1:T} \sqrt{2 \pi}} e^{-\frac{1}{2}\left(\frac{y_{T_l+1:T}-\mu_{T_l+1:T}}{\sigma_{T_l+1:T}}\right)^{2}}
	\end{split}
\end{equation}
where $\mu_{T_l+1:T}$ and $\sigma_{T_l+1:T}$ are mean and standard deviation of teacher predicted distribution. 
The weight $\omega_e(\cdot)$ ranges from 0 to 1 and changes during training. A higher weight signifies a more accurate teacher prediction, optimizing the student network's approximation to the teacher's outputs.

\subsection{Hidden State-driven Online Knowledge Distillation}

\begin{figure}[!ht]
	\centering	  
	\subfigure[]{\includegraphics[width=.45\columnwidth]{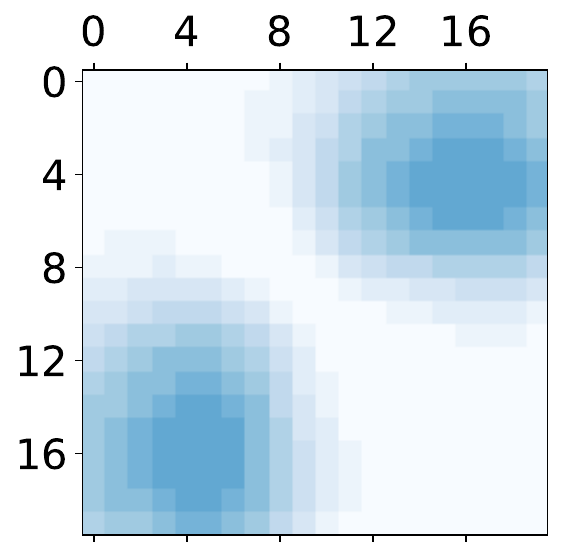}}		\subfigure[]{\includegraphics[width=.45\columnwidth]{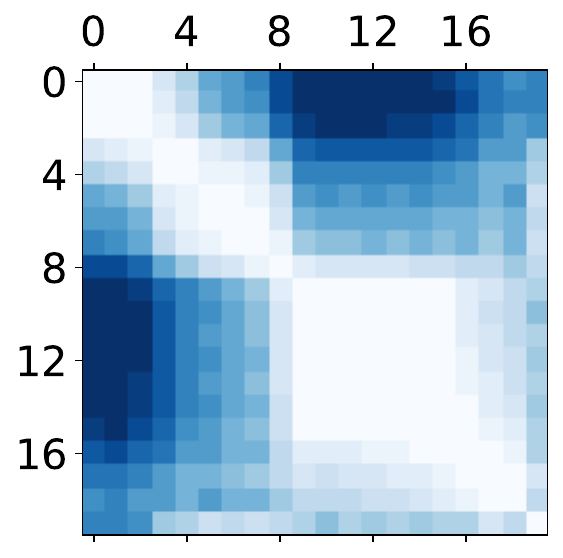}}
	\subfigure[]{\includegraphics[width=.45\columnwidth]{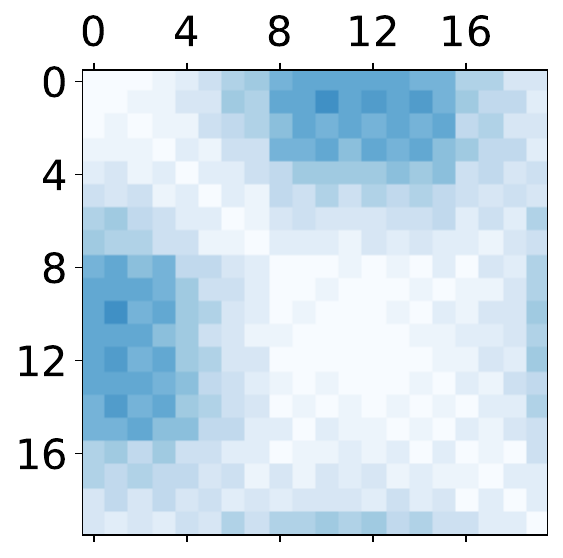}}	
	\subfigure[]{\includegraphics[width=.45\columnwidth]{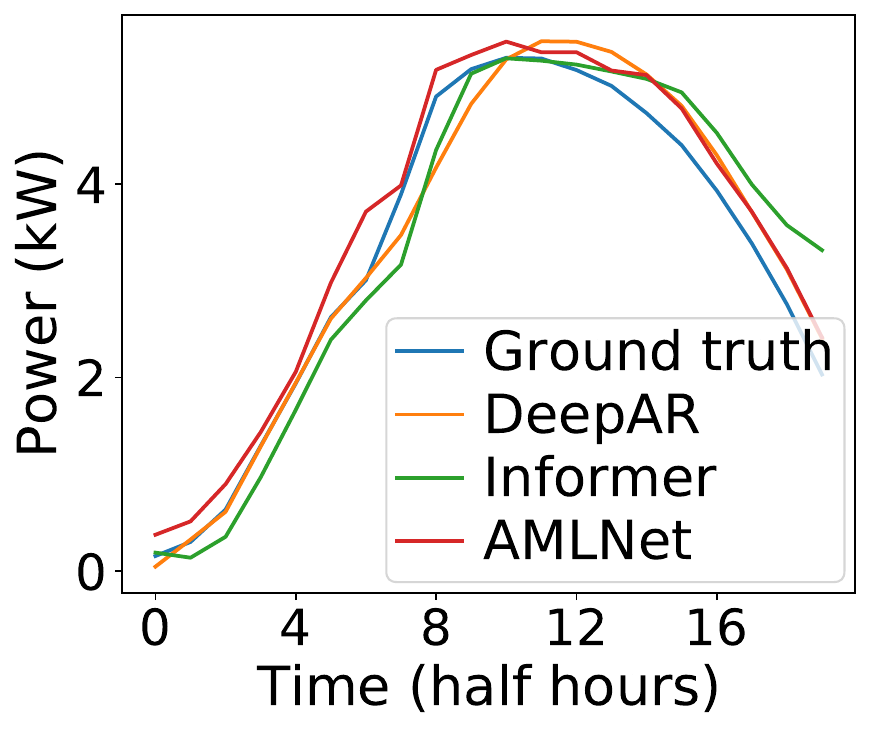}}		
	\caption{Hidden state cosine distance of: (a) DeepAR; (b) Informer and (c) AMLNet. (d) Ground truth vs predictions.}
	\label{Wdis}
\end{figure}

Time series data exhibit a natural continuity, yet NAR forecasting models often yield discontinuous and unrelated predictions \cite{Taieb16}. This disconnect arises due to these models ignoring output interdependence and position information, and we trace this phenomenon to the behavior of hidden states. To illustrate this, we conducted a case study where an AR model outperformed an NAR model. In Figure \ref{Wdis} (d), we display a smooth and continuous PV power measurement from the Sanyo dataset alongside forecasts from an AR model (DeepAR), an NAR model (Informer), and our proposed AMLNet. Notably, the trajectory of DeepAR is noticeably more continuous and smooth than that of the NAR models. To quantify this observation, we employed the Dynamic Time Warping (DTW) algorithm \cite{DTW1978} to measure the similarity between two series. Specifically, we computed the DTW distance and Mean Absolute Percentage Errors (MAPEs) between the DeepAR prediction and the ground truth, yielding 3.05 and 0.102 kW, respectively. In contrast, the Informer yields MAPEs and DTW values of 4.91 and 0.143 kW, indicating that the DeepAR forecasts align more closely with the ground truth.

We visualize the cosine distances between hidden states at the last hidden layer in Figure \ref{Wdis}, where each pixel at row $i$ and column $j$ represents the distance between hidden states $h_i$ and $h_j$ at steps $i$ and $j$. Lighter colors correspond to lower distances. Given two hidden states $h_i$ and $h_j$, their cosine distance is calculated as $1-\frac{h_i \cdot h_j}{||h_i|| \cdot ||h_j||}$. Notably, the average cosine distance between DeepAR hidden states $h_i$ and their six closest neighbors is 0.016, while the corresponding value for Informer is 0.084. Clearly, the cosine distances of DeepAR hidden states are substantially lower, indicating greater similarity to their neighbors. Such similarity suggests that the hidden states of DeepAR generate predictions with consistent and gradual variations, yielding predictions that are more continuous and smooth.
From our analysis, several key observations emerge:
\begin{itemize}
	\item AR models exhibit similar hidden states patterns, while NAR models lack this property.
	\item Dissimilar hidden states could lead to discontinuous predicted trajectory. 
	\item Hidden states hold meaningful information, and distilling knowledge from them could be beneficial \cite{AMLN20ECCV,Chung20ICML}.
\end{itemize}

Thus, we capitalize on the hidden state patterns through online KD to generate continuous predictions while utilizing their inherent information. Unlike offline KD methods which regularize the distance between pairs of hidden states \cite{Hint19EMNLP}, the hidden states of online KD models exhibit greater variability compared to model outputs during training \cite{Chung20ICML}. Direct regularization in this context impairs the learning process and fails to guarantee convergence. To address this, we adopt an adversarial training approach to extract distributional information from hidden states.

In our approach, Peer 1 (P1) learns from Peer 2 (P2) to counter error accumulation, while P2 learns output interdependence from P1. The Student (S) inherits abilities from both P1 and P2. Given a shared encoder, we focus solely on the decoders' hidden states. The adversarial training involves two components: a generator producing feature mappings and a discriminator classifying these mappings. Each decoder layer serves as a generator for feature mappings, and each P1 and P2 layer is paired with a discriminator acting as a classifier (refer to Figure \ref{AMLNet}).

The discriminators receive hidden states $h\in \Re^{T_h \times d_{hid}}$ and output probabilities ranging between 0 (fake) and 1 (real). A discriminator's architecture consists of a sequence of ConvLayer-BatchNorm-LeakyReLU-ConvLayer-LinearLayer-Sigmoid operations. The initial ConvLayer has an output dimension of 16, stride of 2, and kernel size of 3, while the second ConvLayer has an output dimension of 1, stride of 1, and kernel size of 3.

In our training process, the generators aim to fool discriminators by approximating the distribution of hidden states from their teacher networks. Discriminators, on the other hand, strive to distinguish the origin of hidden states. Specifically, we denote the $i$th layer of P1, P2, and S decoders as generators $G_{i,P1}$, $G_{i,P2}$, and $G_{i,S}$, which produce hidden states $h_{i,P1}$, $h_{i,P2}$, and $h_{i,S}$, respectively. The $i$th P1 discriminator $D_{i,P1}$ is trained to classify P1-generated mappings as real (output: 1) and P2 or S-generated mappings as fake (output: 0). Analogously, the $i$th P2 discriminator $D_{i,P2}$ distinguishes P2-generated mappings as real and P1 or S-generated mappings as fake.
Its parameters $G_{i,P1}$ is optimised by minimising the hidden states-driven KD loss $\mathcal{L}_{i,P1}(h_{i,P1})$: 
\begin{equation}
	\begin{split}
		\mathcal{L}_{i,P1}=\alpha_h
		\log (1- D_{i,P2}(h_{i,P1})).
	\end{split}
	\label{LGdAR}
\end{equation}
where hyperparameter $\alpha_h$ controls the weight of hint-driven KD loss.	Similar, $i$th P2 layer $G_{i,P2}$ minimises the KD loss:
\begin{equation}
	\begin{split}
		\mathcal{L}_{i,P2}=\alpha_h
		\log (1- D_{i,P1}(h_{i,P2})).
	\end{split}
	\label{LGdS}
\end{equation}

For hint-driven KD, the student network's shallow layers imitate low-level features from teacher networks, while deep layers learn higher-level features. In the context of the S decoder with fewer layers, we enable each shallow S layer to acquire knowledge from multiple deep network layers. More specifically, the $i$th shallow S network layer learns features from the $j=[1+(i-1)\times\floor{\frac{n_{d}-1}{n_{S}-1}},min(\ceil{\frac{n_{d}-1}{n_{S}-1}}+(i-1)\times\floor{\frac{n_{d}-1}{n_{S}-1}},n_{d})]$th layers of the teachers. Notably, the term $min(\ceil{\frac{n_{d}-1}{n_{S}-1}}+(i-1)\times\floor{\frac{n_{d}-1}{n_{S}-1}},n_{d})$ ensures that $j$ does not exceed the depth of the teacher network.

Consequently, each $i$th shallow S network layer ($G_{i,S}$) distills knowledge from both P1 and P2 decoders, attempting to deceive corresponding discriminators ($D_{j,P1}$ and $D_{j,P2}$) for $j$ within the specified range. The KD loss for the $i$th S layer is formulated in Eq. (\ref{LGS}). Here, $\alpha_h$ governs the weight of hint-driven KD loss.

\begin{equation}
	\begin{split}
		\mathcal{L}_{i,S}=&\alpha_h
		\sum_{j} (\log (1- D_{j,P1}(h_{i,S}))\\
		&+\log (1- D_{j,P2}(h_{i,S}))).
	\end{split}
	\label{LGS}
\end{equation}

In parallel, the discriminators are trained to classify the origin of hidden states. The $i$th P1 discriminator ($D_{i,P1}$) distinguishes features generated by the $i$th P1 layer as real (output: 1), and P2 or S-generated features as fake (output: 0). Similarly, the $i$th P2 discriminator ($D_{i,P2}$) discriminates P2-generated features as real and P1 or S-generated features as fake. The loss functions for the P1 and P2 discriminators are defined in Eq. (\ref{LP1}) and (\ref{LP2}), respectively. Here, $k$ represents the indexes of shallow NAR layers endeavoring to deceive the $i$th discriminator. For example, the first and second S layers aim to deceive the first P1 discriminator in Figure \ref{AMLNet}, resulting in $D_{1,P1}$ having $k={1,2}$.

\begin{equation}
	\begin{split}
		\mathcal{L_D}_{i,P1}=&-\log D_{i,P1}(h_{i,P1})
		\\&-\log (1-D_{i,P1}(h_{i,P2}))
		\\&-\sum_{k}\log (1-D_{i,P1}(h_{k,S}))
	\end{split}
	\label{LP1}
\end{equation}
\begin{equation}
	\begin{split}
		\mathcal{L_D}_{i,P2}=&-\log D_{i,P2}(h_{i,P2})
		\\&-\log (1-D_{i,P2}(h_{i,P1}))
		\\&-\sum_{k}\log (1-D_{i,P2}(h_{k,S}))
	\end{split}
	\label{LP2}
\end{equation}
\subsection{Optimisation and Inference}
The optimization process for AMLNet involves minimizing the forecasting loss, as well as the outcome-driven and hint-driven KD losses. During each training iteration, the encoder and P1 and P2 decoders are optimized first by minimizing Equations (\ref{LossP1}) and (\ref{LossP2}). Subsequently, the S decoder is optimized by minimizing Equation (\ref{LossS}). Lastly, each discriminator associated with the $i$th P1 or P2 layer is optimized by minimizing Equation (\ref{LP1}) or (\ref{LP2}). During testing, only the encoder and S decoder are utilized to generate results. The entire training and testing process is outlined in Algorithm \ref{ADMLlearning}.

\begin{equation}
	\mathcal{L}_{P1}=\mathcal{L_{NLL}}_{P1}+\mathcal{L}_{P1}+\sum_{i=1}^{n_P}\mathcal{L}_{i,P1}
	\label{LossP1}
\end{equation}
\begin{equation}
	\mathcal{L}_{P2}=\mathcal{L_{NLL}}_{P2}+\mathcal{L}_{P2}+\sum_{i=1}^{n_P}\mathcal{L}_{i,P2}
	\label{LossP2}
\end{equation}
\begin{equation}
	\mathcal{L}_{S}=\mathcal{L_{NLL}}_{S}+\mathcal{L}_{S}+\sum_{i=1}^{n_{S}}\mathcal{L}_{i,S}
	\label{LossS}
\end{equation}

\begin{algorithm}
	\caption{Training and Testing Process of AMLNet}
	\begin{algorithmic}[1]
		
		\REQUIRE  Training data $\{(y_{i,1:T-1},x_{i,1:T},y_{i,T_{l+1}:T})\}^{n}_{i=1}$; 
		initialised parameters of the shared encoder $f_{\theta_e}$, P1 decoder $f_{\theta_{P1}}$, P2 decoder $f_{\theta_{P2}}$, shallow S decoder $f_{\theta_{S}}$ and discriminators $\{D_{i,P1}\}^{n_P}_{i=1}$ and $\{D_{i,P2}\}^{n_P}_{i=1}$; training epochs $e_{max}$.
		
		\STATE \textit{// Training stage}
		\FOR{$1 \rightarrow e_{max}$:}
		\STATE \textit{// Optimizing encoder and P1, P2 decoders }
		\STATE Compute encoder output $h_{e;1:T_l}$ (Eq. (\ref{theta_e}))
		\STATE Compute P1 and P2 decoders forecasts $y_{P1;T_l+1:T}$ and $y_{P2;T_l+1:T}$ and their hidden states $h_{1:n_P,P1}$ and $h_{1:n_P,P2}$ (Eq. (\ref{theta_P1} and \ref{theta_P2}))

		\STATE Compute loss with P1 forecasts $\mathcal{L}_{P1}$ (Eq. (\ref{LossP1})) 
		\STATE Compute loss with P2 forecasts $\mathcal{L}_{P2}$ (Eq. (\ref{LossP2})) 
		\STATE Update encoder, P1, P2 decoders by minimising $\mathcal{L}_{P1}$ and $\mathcal{L}_{P2}$
		
		\STATE \textit{// Optimizing S decoder}
		\STATE Compute forecasts $y_{S;T_l+1:T}$ and hidden states $h_{1:n_{S},S}$ of shallow S decoder (Eq. (\ref{theta_S})) 
		\STATE Compute loss with S forecasts $\mathcal{L}_{S}$ (Eq. (\ref{LossS})) 
		\STATE Update  S decoder by minimising $\mathcal{L}_{S}$ 
		
		\STATE \textit{// Optimizing discriminators}
		\STATE Compute classification loss $\mathcal{L_D}_{i,P1}$ and $\mathcal{L_D}_{i,P2}$ for every P1 and P2 discriminators (Eq. (\ref{LP1} and \ref{LP2})) 
		\STATE Update discriminators by minimising the classification loss 
		
		\ENDFOR
		\STATE \textit{// Testing stage}
		\STATE Compute encoder output $h_{e;1:T_l}$ (Eq. (\ref{theta_e}))
		\STATE Compute forecasting results $y_{S;T_l+1:T}$ by the S decoder (Eq. (\ref{theta_S}))
	\end{algorithmic}
	\label{ADMLlearning}
\end{algorithm}

\section{Experiments}

\subsection{Experimental Details}
We compare the performance of AMLNet with six methods: four state-of-the-art deep learning (DeepAR, LogSparse Transformer, N-BEATS and Informer), a statistical (SARIMAX) and a persistence model. 
1) \textbf{Persistence} is a typical baseline in forecasting which considers the time series of the previous day as the prediction for the next day; 
2) \textbf{SARIMAX} \cite{Durbin01book} is an extension of ARIMA which cann handle seasonality with exogenous variables; 
3) \textbf{DeepAR} \cite{DeepAR20} is a widely used RNN-based forecasting model; 
4) \textbf{LogSparse Transformer} \cite{Logsparse19NIPS} is a Transformer-based forecasting model, it is denoted as "LogTrans" in Table \ref{Accuracy}; 
5) \textbf{N-BEATS} \cite{N-BEATS} consists of blocks of fully-connected neural networks, organised into stacks using residual links. We introduced covariates at the input of each block to facilitate multivariate series forecasting;
6) \textbf{Informer} \cite{Informer21} is a Transformer-based forecasting model. 
We modified it for probabilistic forecasts to generate the mean value and variance. 
Note that Persistence, N-BEATS and Informer are NAR models while the others are AR models.

\begin{table*}[!t]
	\renewcommand{\arraystretch}{1}	
	\centering
	\begin{tabular}{C{1.5cm} C{.8cm} C{.8cm} C{.8cm} C{.8cm} C{.8cm} C{.8cm} C{.8cm} C{.8cm} C{1cm} C{.8cm} C{.8cm} } 
		\specialrule{.1em}{.05em}{.05em} 
		&$\lambda_G$&$\lambda_D$&$\alpha_o$&$\alpha_h$& $\delta$ & $d_{hid}$ & $n_{e}$& $n_{d}$&$n_{S}$ & $d_{f}$ &$n_{h}$\\	
		\hline
		Sanyo&0.005&0.001&0.1&0.5& 0 & 48 &4& 4 &2 &16&8\\				
		Hanergy&0.005&0.001&0.1&0.5&0 & 48 & 4&4&2&16 &8 \\			
		Solar&0.005&0.001&0.5&0.001& 0.2 & 96 & 4&3&2 &48 &32\\			
		Electricity&0.001&0.001&0.1&0.001& 0.1 & 48 &4& 3&2 &48 &32\\		
		\specialrule{.1em}{.05em}{.05em} 
	\end{tabular}
	\caption{Hyperparameters for AMLNet}
	\label{AMLNet_Parameters} 
\end{table*}

All models were implemented using PyTorch 1.6 and evaluated on Tesla V100 16GB GPU.
The deep learning models were optimised by mini-batch gradient descent with the Adam optimiser and a maximum number of epochs 200. 
Following the experimental setup in \cite{Yang20ICONIP,9679135}, we used the following training, validation and test split: 	for Sanyo and Hanergy - the data from the last year as test set, the second last year as validation set for early stopping and the remaining data (5 years for Sanyo and 4 years for Hanergy) as training set; for Solar and Electricity - the last week data as test set (from 25/08/2006 for Solar and 01/09/2014 for Electricity), the week before as validation set. 
For all data sets, the data preceding the validation set is split in the same way into three subsets and the corresponding validation set is used to select the best hyperparameters.
We selected the hyperparameters with a minimum loss on the validation set. 
We used Bayesian optimisation for hyperparameter search with a maximum number of iterations 20. The models used for comparison were tuned based on the authors' recommendations. 
For the Transformer-based models, we used learnable position and ID (for Solar, Electricity and Exchange sets) embedding. 
For AMLNet, the constant sampling factor for Informer backbone was set to 2, and the length of start token $T_de$ is fixed as half of the forecasting horizon. The learning rate of generator $\lambda_G$ and discriminator  $\lambda_D$ was fixed, the loss function regularisation parameters $\alpha_o$ and $\alpha_h$ were chosen from \{0.001, 0.05, 0.1, 0.5\}, the dropout rate $\delta$ was chosen from \{0, 0.1, 0.2\}, the hidden layer dimension size $d_{hid}$ was chosen from \{8, 12, 16, 24, 48\}, the Informer backbone Pos-wise FFN dimension dimension size $d_{f}$ and number of heads $n_{h}$ were chosen from \{8, 12, 16, 24, 48, 96\} and \{4, 8, 16, 24, 32\}, number of encoder $n_e$, P1 and P2 decoder $n_{S}$ and shallow NAR decoder layers $n_{S}$ were chosen from and \{2, 3, 4\}. Note that number of encoder layer is not less than number of decoder layers, P1 and P2 decoders have same number of layers, shallow NAR decoder has less number of layers than deep decoders.  	
The discriminators are simply a series of ConvLayer-BatchNorm-LeakyReLU-ConvLayer-LinearLayer-Sigmoid. The first ConvLayer has the output dimension of 16, stride of 2 and kernel size of 3, while the second has dimension of 1, stride of 1 and kernel size of 3. 

The selected best hyperparameters for AMLNet are listed in Table \ref{AMLNet_Parameters} and used for the evaluation of the test set.

Following \cite{DeepAR20}, we report the standard $\rho$0.5 and $\rho$0.9-quantile losses. 
The quantile loss function is applied to predict quantiles, and quantile $\rho$ is the value below which a fraction of observations in a group falls. 
Given the ground truth $y$ and $\rho$-quantile of the predicted distribution $\hat{y}$, the $\rho$-quantile loss is given by $\mathrm{QL}_{\rho}(y, \hat{y})$:
\begin{equation}
	\begin{split}
		\mathrm{QL}_{\rho}(y, \hat{y})&=\frac{2\times\sum_{t} P_{\rho}\left(y_{t}, \hat{y}_{t}\right)}{\sum_{t}\left|y_{t}\right|}, 
		\\
		\quad P_{\rho}(y, \hat{y})&=\left\{\begin{array}{ll}
			\rho(y-\hat{y}) & \text { if } y>\hat{y} \\
			(1-\rho)(\hat{y}-y) & \text { otherwise }
		\end{array}\right.
	\end{split}
\end{equation}

\begin{table*}[!ht]
	\renewcommand{\arraystretch}{1}	
	\centering
	\begin{tabular}{ C{2cm}   C{2.2cm}  C{2.2cm}  C{2.2cm}  C{2.2cm}  } 
		\specialrule{.1em}{.05em}{.05em} 
		& Sanyo& Hanery& Solar& Electricity\\
		\hline
		Persistence &0.154/-&0.242/-&0.256/-&0.091/-\\ 
		SARIMAX&0.124/0.096&0.145/0.098&0.256/0.192&0.196/0.079\\ 
		DeepAR 
		&0.070/0.031&0.092/0.045&0.222$^\diamond$/0.093$^\diamond$&0.075$^\diamond$/0.040$^\diamond$\\		
		LogTrans
		&0.067/0.036&0.124/0.066&0.210$^\diamond$/0.082$^\diamond$&\textbf{0.059}$^\diamond$/{0.034}$^\diamond$\\
		N-BEATS&0.077/-&0.132/-&0.212/-&0.071/-\\			
		Informer&{0.046}/{0.022}&{0.084}/{0.046}&{0.215}/{0.115}&0.068/0.033\\
		\hline
		AMLNet-P1&0.044/{0.021}&0.084/0.043&0.224/0.091&0.068/0.034\\
		AMLNet-P2&\textbf{0.040}/\textbf{0.019}&0.078/0.040&0.206/0.090&0.065/0.033\\
		AMLNet-S&{0.042}/0.020&\textbf{0.077}/\textbf{0.038}&\textbf{0.204}/\textbf{0.088}&{0.067}/\textbf{0.032}\\	
		\specialrule{.1em}{.05em}{.05em} 
	\end{tabular}
	\caption{$\rho$0.5/$\rho$0.9-loss of data sets with various granularities. $\diamond$ denotes results from \protect\cite{Logsparse19NIPS}.}
	\label{Accuracy} 
\end{table*}

\subsection{Accuracy Analysis}

Table \ref{Accuracy} shows the $\rho$0.5 and $\rho$0.9 losses of all models, including the three AMLNet versions which use different decoders (AMLNet-P1, AMLNet-P2, AMLNet-S). As N-BEATS and Persistence produce point forecasts, only the $\rho$0.5-loss is reported for them. 
AMLNet is the most accurate method - it outperforms the other methods on all data sets except for $\rho$0.5 on Electricity where the Logsparse Transformer is the best model. 
For AMLNet, S decoder successfully inherits the abilities of P1 and P2 decoders with fewer layers and has the best performance on Hanergy and Solar sets.
Overall, all decoders of AMLNet outperform their backbone model Informer except for Solar and Electricity sets where P1 underperforms, indicating the AMLNet is beneficial. Both NAR branches (P2 and S) exhibit clear improvement over Informer, 
suggesting the design for overcoming the disadvantages of AR and NAR models is successful for improving forecasting accuracy. 

AST \cite{AST20NIPS} employs adversarial training to improve the contiguous and fidelity at the sequence level. AST is not compared directly because it generates quantile forecasts and minimises quantile loss, while we consider probabilistic forecasting with different objective functions in Eq. (\ref{NLL}). To compare the effectiveness of AMLNet with AST, we apply the adversarial training of AST to AMLNet and its backbone - Informer, and the results are shown in Table \ref{Accuracy_AST}. The adversarial training improves the performance of Informer, while  AMLNet still exhibits advantages. The design of AMLNet is compatible with the AST adversarial training, combining both techniques achieves further performance improvement on the S decoder.

\begin{table}[!ht]
	\renewcommand{\arraystretch}{1}	
	\centering
	\begin{tabular}{ C{2.6cm}   C{2.2cm}  C{2.2cm}    } 
		\specialrule{.1em}{.05em}{.05em} 
		& Sanyo& Hanery\\
		\hline		
		Informer&{0.046}/{0.022}&{0.084}/{0.046}\\
		\hline		
		Informer+Adv&{0.045}/{0.022}&{0.079}/{0.041}\\
		\hline
		AMLNet-P1&0.044/{0.021}&0.084/0.043\\
		AMLNet-P2&{0.040}/\textbf{0.019}&0.078/0.040\\
		AMLNet-S&{0.042}/0.020&{0.077}/{0.038}\\
		\hline
		AMLNet-P2+Adv&{0.047}/{0.023}&{0.079}/{0.039}\\	
		AMLNet-P1+Adv&{0.045}/{0.022}&{0.099}/{0.047}\\	
		AMLNet-S+Adv&\textbf{0.039}/{0.020}&\textbf{0.072}/\textbf{0.035}\\	
		\specialrule{.1em}{.05em}{.05em} 
	\end{tabular}
	\caption{$\rho$0.5/$\rho$0.9-loss of adversarial training study.}
	\label{Accuracy_AST} 
\end{table}

\subsection{Case Analysis}

To study the capabilities of AMLNet in addressing error accumulation and modeling output space interdependence, we conduct a comparative analysis with two benchmark models:  classic AR model DeepAR and  NAR model Informer. The evaluation is performed on both the Sanyo and Hanergy datasets, providing insights into AMLNet's performance across different scenarios.

\begin{figure}[!ht]
	\centering	  
	\subfigure[]{\includegraphics[width=.8\columnwidth]{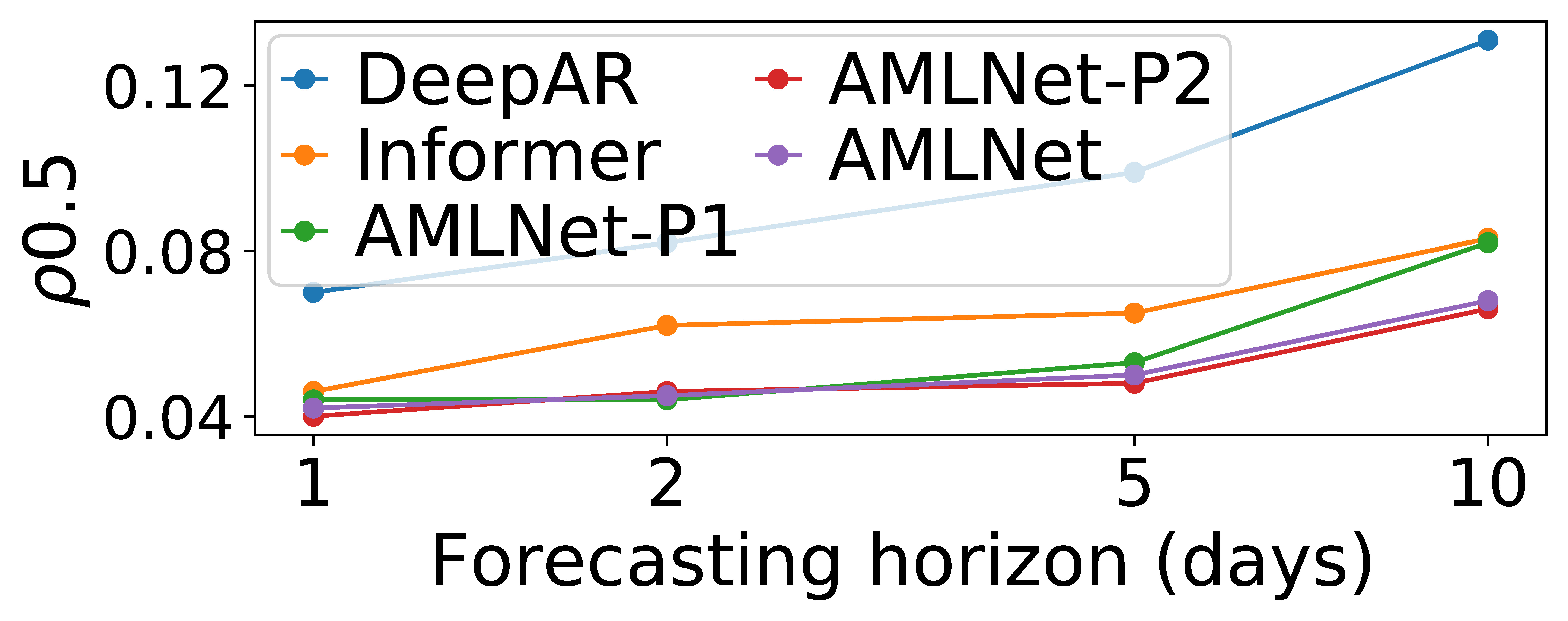}}
	\subfigure[]{\includegraphics[width=.8\columnwidth]{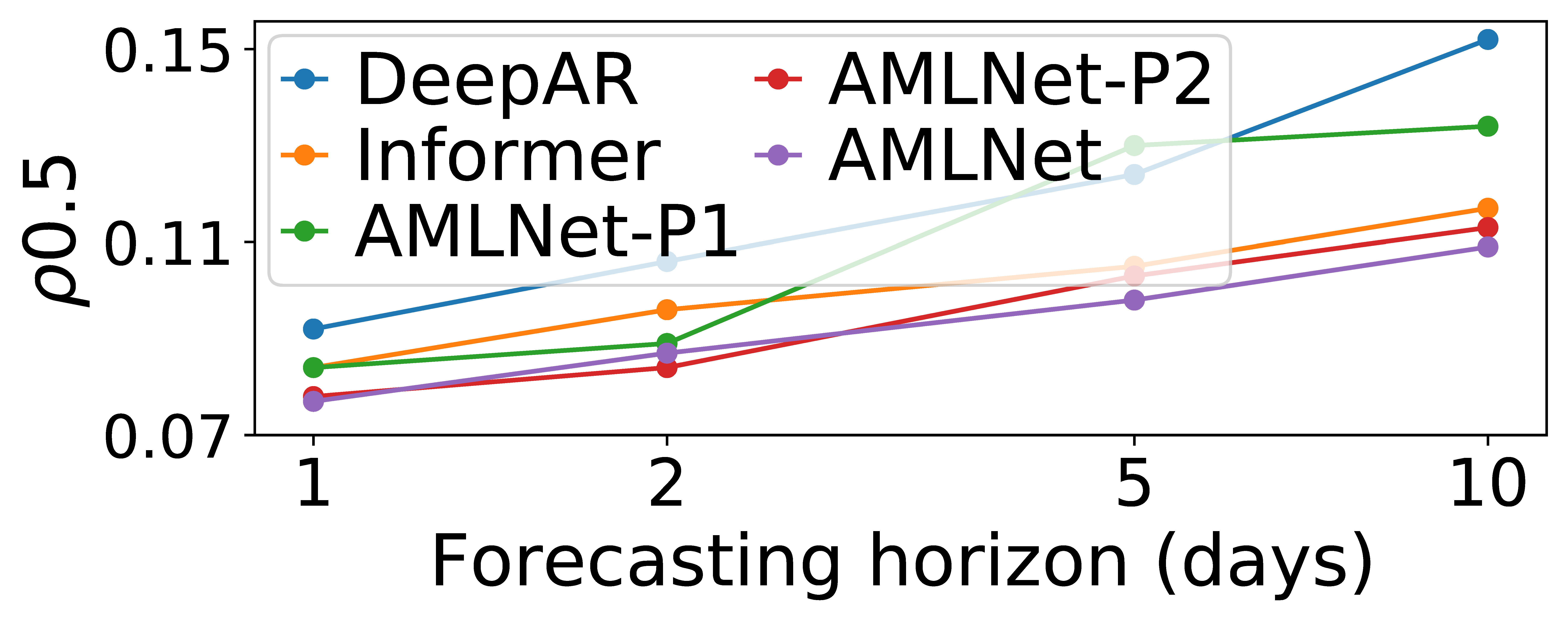}}	
	\caption{$\rho$0.5-loss of DeepAR, Informer and AMLNet with various forecasting horizon on (a) Sanyo set and (b) Hanergy set}
	\label{Error_multiH_full}
\end{figure}

Fig. \ref{Error_multiH_full} illustrates the $\rho$0.5-loss of various models across different forecasting horizons, using a fixed pasting history. The loss of all models tends to increase as the forecasting horizon expands. However, it is evident that the performance of AR models, such as DeepAR, deteriorates more significantly compared to NAR models. Remarkably, AMLNet's P1 decoder consistently outperforms DeepAR across different horizons, demonstrating its capability to mitigate the adverse effects of error accumulation. Conversely, NAR models, including Informer and AMLNet, exhibit relatively stable performance over varying forecasting horizons. This observation indicates that AMLNet's design effectively addresses the issue of error accumulation in its P1 decoder.

Referencing Fig. \ref{Wdis} (c), it becomes evident that AMLNet's S decoder exhibits lower cosine distances in hidden states compared to its backbone counterpart shown in Fig. \ref{Wdis} (b). This distinction is particularly pronounced when observing the lighter color in Fig. \ref{Wdis} (c) as compared to Fig. \ref{Wdis} (b). Additionally, the average cosine distances of the hidden states in the P2 and S decoders on both the Sanyo and Hanergy datasets are significantly lower, by 28\% and 23\% respectively, compared to the backbone model. Furthermore, the average DTW distances of the P2 and S decoder predictions exhibit a reduction of 18\% and 17\%, respectively. These findings underline the efficacy of our designed approach in learning and leveraging output space interdependence. This enables the model's hidden states to exhibit greater similarity to neighboring states and subsequently generates more realistic and coherent prediction trajectories.

	\begin{table*}[!ht]
		\renewcommand{\arraystretch}{1}	
		\centering
		\begin{tabular}{ C{2cm}   C{2cm}  C{2cm}  C{2cm}  C{2cm}  } 
			\specialrule{.1em}{.05em}{.05em} 
			& Sanyo& Hanery& Solar& Electricity\\
			\hline
			LogTrans&101.5$\pm$3.5&112.7$\pm$7.7&171.8$\pm$9.6&437.4$\pm$21.5\\
			Informer&18.1$\pm$0.5&18.7$\pm$1.1&44.7$\pm$0.6&213.7$\pm$0.6\\
			AMLNet-P1&150.2$\pm$2.4&148.4$\pm$0.8&249.5$\pm$0.4&600.1$\pm$0.7\\
			AMLNet-P2&21.7$\pm$0.2&21.5$\pm$0.1&48.4$\pm$0.1&289.6$\pm$0.4\\
			AMLNet-S&11.4$\pm$0.5&11.6$\pm$0.5&30.0$\pm$5.8&152.2$\pm$0.2\\
			\specialrule{.1em}{.05em}{.05em} 
		\end{tabular}
		\caption{Inference time (ms) of data sets.}
		\label{Inference_time} 
	\end{table*}

	\subsection{Speed Analysis}
We conducted an evaluation of the inference time for different configurations of AMLNet, as well as the NAR backbone Informer and the AR baseline LogTrans. The results are summarized in Table \ref{Inference_time}. All experiments were conducted on the same computer configuration, and the reported values represent the average elapsed time in milliseconds along with the standard deviation from 10 runs. Notably, the NAR models exhibit faster inference times compared to the P1 decoder, primarily due to their inherent parallelizability. Informer and AMLNet-P2 demonstrate similar inference speeds, which is consistent with their comparable architectural characteristics. AMLNet-S, designed with fewer layers, stands out as the fastest among the models evaluated.

	\subsection{Ablation Analysis}
To assess the effectiveness of our proposed methods, we conducted an ablation study, focusing on the $\rho$0.5-loss metric for various model configurations. The results are presented in Table \ref{Ablation_Accuracy_full}. In this table, $\mathcal{L}_{o}$ corresponds to the classic online KD, $\mathcal{L}_{wo}$ represents our outcome-driven KD, $\mathcal{L}_{GAN}$ indicates adversarial KD applied to the last hidden layer, and $\mathcal{L}_{hGAN}$ refers to our hint-driven KD.

Among the key findings from this ablation analysis:
\begin{itemize}
	\item AMLNet, when combined with our proposed KD methods, emerges as the most effective model configuration, attaining the highest accuracy.
	\item Both  outcome-driven KD and hint-driven KD  lead to improved accuracy when incorporated into the frameworks, underscoring the efficacy of both design approaches.
	\item The  Backbone+$\mathcal{L}_{wo}$+$\mathcal{L}_{GAN}$ outperforms Backbone+$\mathcal{L}_{o}$+$\mathcal{L}_{hGAN}$ substantially, suggesting  outcome-driven KD exerts a more pronounced impact on accuracy enhancement compared to the hint-driven KD.
	\item  S decoders tend to outperform P1 and P2 decoders, supporting the notion that our design of online knowledge distillation from P1 and S is a beneficial strategy.
\end{itemize}

	\begin{table}[!t]
		\renewcommand{\arraystretch}{1}	
		\centering
				\begin{tabular}{ C{1.9cm} C{.4cm}   C{1.4cm}  C{1.4cm}  C{1.4cm}    } 
			\specialrule{.1em}{.05em}{.05em} 
			&& Sanyo& Hanery& Solar \\
			\hline
			\multicolumn{2}{c}{Backbone (Informer)}
			&{0.046}/{0.022}&{0.084}/{0.046}&{0.215}/{0.115}\\
			\hline
			\multirow{3}{*}{\shortstack[c]{Backbone\\+$\mathcal{L}_{o}$\\(DML)}}
			&P1&0.053/0.024&0.098/0.048&0.258/0.090\\
			&P2&0.051/0.024&0.092/0.047&0.211/0.104\\
			&S&0.048/0.023&0.083/0.042&0.219/0.111\\	
			
			\hline
			\multirow{3}{*}{\shortstack[c]{Backbone\\+$\mathcal{L}_{o}$+$\mathcal{L}_{GAN}$\\(AFD)}}
			&P1&0.053/0.023&0.096/0.046&0.556/0.202\\
			&P2&0.043/0.020&0.079/0.040&0.228/0.104\\
			&S&0.048/0.022&0.084/0.041&0.221/0.100\\	
			
			\hline
			\multirow{3}{*}{\shortstack[c]{Backbone\\+$\mathcal{L}_{o}$+$\mathcal{L}_{hGAN}$}}
			&P1&0.051/0.023&0.095/0.046&0.276/0.098\\
			&P2&0.045/0.021&0.095/0.045&0.266/0.150\\
			&S&0.046/0.022&0.079/0.039&0.233/0.108\\	
			
			\hline
			\multirow{3}{*}{\shortstack[c]{Backbone\\+$\mathcal{L}_{wo}$}}
			&P1&0.049/0.022&0.087/0.045&0.235/0.089\\
			&P2&0.045/0.022&0.093/0.047&0.215/0.090\\
			&S&0.049/0.023&0.083/0.041&0.208/0.086\\ 
			
			\hline
			\multirow{3}{*}{\shortstack[c]{Backbone\\+$\mathcal{L}_{wo}$+$\mathcal{L}_{GAN}$}}
			&P1&0.043/0.020&0.082/0.040&0.215/0.086\\ 
			&P2&0.043/0.020&0.079/0.038&0.205/\textbf{0.087}\\ 
			&S&0.043/0.020&0.079/\textbf{0.037}&0.205/0.089\\ 
			
			\hline
			\multirow{3}{*}{\shortstack[c]{Backbone\\+$\mathcal{L}_{wo}$+$\mathcal{L}_{hGAN}$\\(AMLNet)}}
			&P1&0.044/{0.021}&0.084/0.043&0.224/0.091\\
			&P2&\textbf{0.040}/\textbf{0.019}&0.078/0.040&0.206/0.090\\
			&S&{0.042}/0.020&\textbf{0.077}/{0.038}&\textbf{0.204}/{0.088}\\	
			
			\specialrule{.1em}{.05em}{.05em} 
		\end{tabular}
		\caption{$\rho$0.5/$\rho$0.9-loss of data sets for ablation study.}
		\label{Ablation_Accuracy_full} 
	\end{table}
	
	\section{Conclusion}
	
	We introduce AMLNet, the NAR model that harnesses both outcome-driven and hint-driven online KD methods. It comprises a shared encoder, alongside deep AR (P1), deep NAR (P2), and shallow NAR (S) decoders. P1 and P2 operate collaboratively, mutually distilling knowledge from each other, and collectively acting as ensemble teachers to effectively transfer this knowledge to S.
	Our method dynamically assigns attention-based weights to the model's output KD loss, thereby effectively mitigating the risk of learning from less reliable predictions. Additionally, we employ adversarial training to distill knowledge from the distribution of hidden states. This is particularly significant as the root of unrealistic forecasts in NAR models often lies within the hidden states, which inherently carry valuable information.
	Our extensive experimental evaluations substantiate the remarkable performance and effectiveness of AMLNet in comparison to state-of-the-art forecasting models and existing online KD methods. AMLNet excels not only in modeling output space interdependence, resulting in more plausible forecasts, but also in addressing the challenge of error accumulation, all while maintaining a low inference latency.

	\bibliographystyle{IEEEtran}
	\bibliography{Bibliography/bib_yang}
	
	\appendix
	
	\section{Dynamic Time Warping}
	DTW is a classic algorithm to measure similarity between two trajectories which may have different length and vary in time \cite{DTW1978}. It finds the optimal alignments between two series via dynamic programming. 
	Given two univariate time series $\mathbf{x}\coloneqq[x_1,...,x_n]\in \Re^{1\times n}$ and $\mathbf{y}\coloneqq[y_1,...,y_m]\in \Re^{1\times m}$, DTW can be considered as a optimization problem:
	\begin{equation}
		\begin{split}
			\text{DTW}(\mathbf{x},\mathbf{y})
			=\min\limits_{\pi\in \mathcal{A}(\mathbf{x},\mathbf{y})} \sum_{(i,j)\in\pi}d({x_i},{y_j})
			=\min\limits_{\pi\in \mathcal{A}(\mathbf{x},\mathbf{y})} \sum_{(i,j)\in\pi} |{x_i}-{y_j}|
		\end{split}
		\label{DTW_eq}
	\end{equation}
	where $\mathcal{A}$ is the optimal path between $\mathbf{x}$ and $\mathbf{y}$, $\pi$ is a sequence of time index pairs and $d({x_i},{y_j})=|{x_i}-{y_j}|$ is the distance measure. The alignment $\pi$ increases monotonically and continuously from $(1,1)$ to $(n,m)$.  
	
	The details of DTW is shown in algorithm \ref{DTW} where line 4 involves dynamic programming with a quadratic cost $O(nm)$. 
	\begin{algorithm}
		\caption{Dynamic Time Warping}
		\begin{algorithmic}[1]
			\REQUIRE  $\mathbf{x}\coloneqq[x_1,...,x_n]\in \Re^{1\times n}$, $\mathbf{y}\coloneqq[y_1,...,y_m]\in \Re^{1\times m}$.
			\STATE $d_{0,0}=0$; $d_{i,0}=d_{0,j}=\infty$; $i\in[\![1,N]\!]$, $j\in[\![1,m]\!]$
			\FOR{$j\leftarrow 1$ to $m$:}	
			\FOR{$i\leftarrow 1$ to $n$:}
			\STATE $d_{i,j}=|{x_i}-{y_j}|+\min \{d_{i-1,j-1},d_{i-1,j},d_{i,j-1}\}$
			\ENDFOR		
			\ENDFOR
			\STATE Return $d_{n,m}$
		\end{algorithmic}
		\label{DTW}
	\end{algorithm}

\end{document}